\documentclass[letterpaper]{article} %
\usepackage{aaai2026}  %
\usepackage{times}  %
\usepackage{helvet}  %
\usepackage{courier}  %
\usepackage[hyphens]{url}  %
\usepackage{graphicx} %
\usepackage{natbib}  %
\usepackage{caption} %
\usepackage{algorithm}
\usepackage{algorithmic}

\usepackage{lineno}
\usepackage{newfloat}
\usepackage{listings}
\DeclareCaptionStyle{ruled}{labelfont=normalfont,labelsep=colon,strut=off} %
\floatstyle{ruled}
\newfloat{listing}{tb}{lst}{}
\floatname{listing}{Listing}
\usepackage{amsmath}
\usepackage{booktabs}
\usepackage{multirow}
\usepackage[table, dvipsnames]{xcolor}
\newcommand{\lowbetter}{$_\downarrow$}
\usepackage[T1]{fontenc}
\usepackage{amssymb}

\title{Mitigating Negative Flips via Margin Preserving Training}
\author{
    Simone Ricci,
    Niccolò Biondi,
    Federico Pernici,
    Alberto Del Bimbo
}
\affiliations{
    DINFO (Department of Information Engineering), University of Florence, Italy\\
    MICC (Media Integration and Communication Center)\\

    <name>.<surname>@unifi.it
}

\begin{document}

\maketitle

\begin{abstract}
Minimizing inconsistencies across successive versions of an AI system is as crucial as reducing the overall error.
In image classification, such inconsistencies manifest as negative flips, where an updated model misclassifies test samples that were previously classified correctly.
This issue becomes increasingly pronounced as the number of training classes grows over time, since adding new categories reduces the margin of each class and may introduce conflicting patterns that undermine their learning process, thereby degrading performance on the original subset.
To mitigate negative flips, we propose a novel approach that preserves the margins of the original model while learning an improved one.  
Our method encourages a larger relative margin between the previously learned and newly introduced classes by introducing an explicit margin-calibration term on the logits.
However, overly constraining the logit margin for the new classes can significantly degrade their accuracy compared to a new independently trained model. 
To address this, we integrate a double-source focal distillation loss with the previous model and a new independently trained model, learning an appropriate decision margin from both old and new data, even under a logit margin calibration.
Extensive experiments on image classification benchmarks demonstrate that our approach consistently reduces the negative flip rate with high overall accuracy.
\end{abstract}

\begin{links}
    \link{Code}{https://github.com/miccunifi/negative_flip_MPT}
\end{links}

\section{Introduction}\label{sec:intro}

\begin{figure}[t]
    \centering
     \includegraphics[width=0.98\linewidth]{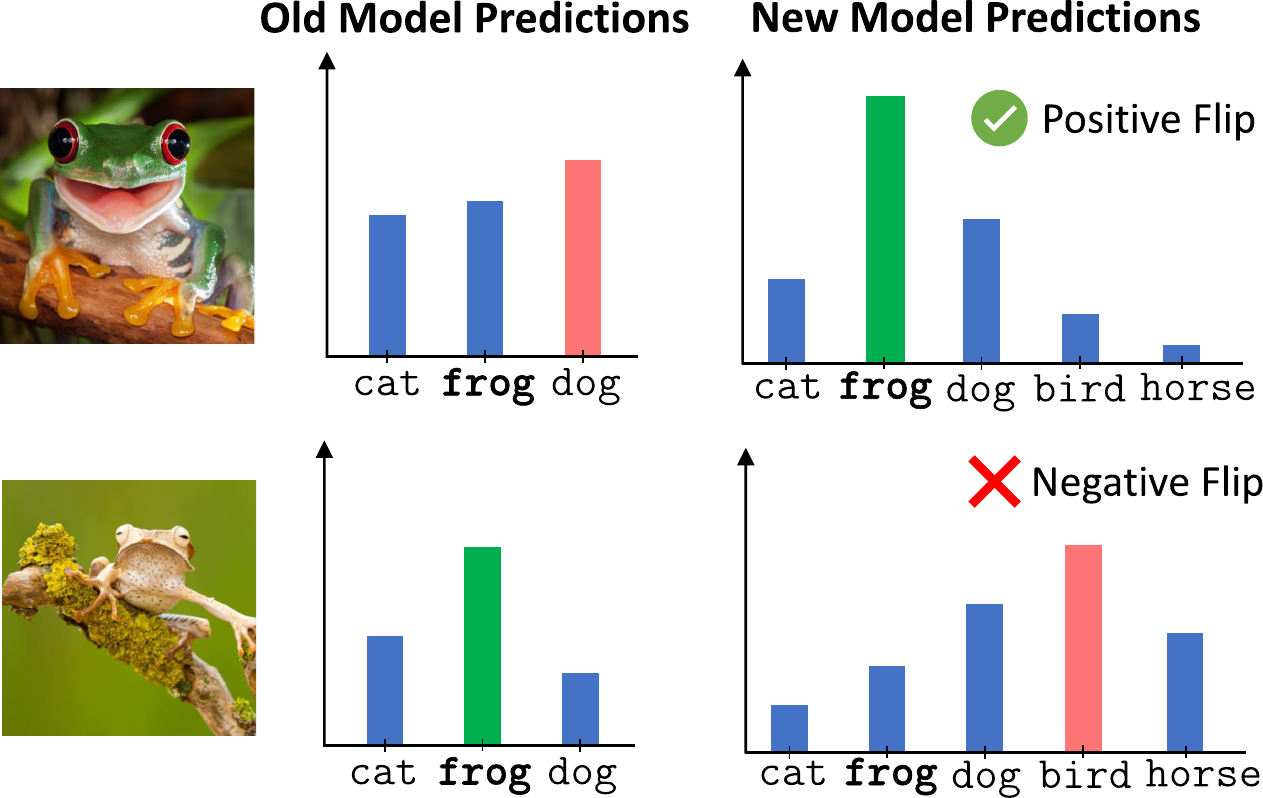}
    \caption{Illustration of possible two prediction changes following a model update. Each row shows the predictions of the old model (left) and the new model (right) for a given image, showing two possible cases: (1) a Positive Flip, where the old model was wrong and the new model is correct---the desirable outcome after a model update; (2) a Negative Flip, where the old model was correct but the new model is wrong, which can lead to unexpected system behaviors. In this paper, we focus on reducing Negative Flips in a classification task.}
    \label{fig:prediction_flip}
\end{figure}

Recent advances in machine learning have required the frequent deployment of updated models in production systems \cite{Raffel2023, yadav2024survey}. These updates often introduce new models that leverage more expressive network architectures \citep{touvron2023llama}, novel training techniques or paradigms \citep{biondi2024stationary, echterhoff2024muscle, shen2020towards, zhou2023bt}, and additional data \citep{gunasekar2023textbooks}.
However, replacing an existing model requires balancing the potential reduction in overall error rates against the risk of introducing new errors that can disrupt downstream processes \cite{milani2016launch, yan2021positive} or lead to unexpected system behaviors for human users \cite{bansal2019updates}. 
This regression\footnote{The term \textit{regression}---from the software industry---denotes a decline in system performance following an update. Although an updated model may demonstrate improved average performance, any instances of regression can potentially disrupt downstream post-processing workflows.} phenomenon, observed in both visual and natural language tasks~\cite{yan2021positive, abdul2024align}, arises from the occurrence of negative flips---instances that are correctly classified by the old model but misclassified by the new one (see Figure \ref{fig:prediction_flip}).

Negative flips in classification tasks are most likely to occur when the number of classes increases to accommodate new categories \cite{yan2021positive, zhao2024elodi, zhang2021hot}. 
From a margin-based perspective, increasing the number of classes at each model update reduces the margin between adjacent classes, due to class means pushing themselves to arrange in a way that maximizes the minimum one-vs-rest margin \cite{pmlr-v235-jiang24i, papyan2020prevalence, yaras2022neural, tirer2023perturbation, pernici2019fix, pernici2021regular}.
In addition to the reduction imposed by the optimization process and geometric constraints \cite{pmlr-v235-jiang24i}, when new categories are introduced, the decision margins are subject to further challenges arising from negative transfer \cite{li2023identification, zhang2022survey}. As the model learns to encode features that discriminate among an expanded set of classes, the incorporation of additional categories may lead to the emergence of interfering features, thereby adversely impacting the performance on previously acquired tasks or classes \cite{tirer2023perturbation, liu2021interclass, torralba2007sharing}. In extreme cases, negative flips involve not only ambiguous samples near the decision boundaries but also high-confidence samples influenced, for example, by spurious correlations \cite{wang2021causal}. 

Despite these challenges, most existing approaches for reducing negative flips \cite{yan2021positive, zhao2024elodi, zhang2021hot} overlook a key factor: \textit{the progressive reduction of class margins as new categories are added}. Smaller margins make the model more susceptible to negative flips, since even confident, previously correct predictions can become incorrect due to slight shifts in decision boundaries. Therefore, preserving or restoring these margins is critical for maintaining consistent performance across model updates—especially for previously learned classes. This observation motivates us to focus on margin preservation as a central component of our approach.
To this end, we preserve the logit decision margins\footnote{The logit decision margin is the difference between the logit of the ground-truth class and the largest logit among all other classes. Following \citet{pleiss2020identifying} notation, the logit margin for an input sample of class~$y$ is defined as $\gamma(\mathbf{z},y) = \mathbf{z}_y - \max_{j \neq y} \mathbf{z}_j$, where $\mathbf{z}$ denotes the logit vector from the final (pre-softmax) layer.} associated with previously learned classes, thereby addressing the challenge of negative flips when new classes are incrementally introduced.
Specifically, we introduce a positive logit bias for the new classes as a margin-calibration term within the softmax cross-entropy loss, thereby modifying the optimization process such that the decision margins are biased towards the previously learned classes.
Although this adjustment effectively reduces negative flips for previously learned classes, it may also result in underfitting of the newly introduced ones, as the model is encouraged to learn a smaller margin for them.
To overcome this limitation and ensure appropriate margin learning for both previously learned and newly introduced classes, we propose a double-source focal distillation strategy.
We distill knowledge corresponding to the correct predictions on the newly introduced categories from a model independently trained on the complete set of classes, while at the same time leveraging the previous model to preserve its correct behavior on the original categories.

Our main contributions are as follows:
\begin{itemize}
\item We present a novel approach, called Margin Preserving Training, to address the reduction of negative flips by focusing on the preservation of decision margins for previously learned classes.
\item We propose a double-source focal distillation strategy to mitigate the negative impact of the margin-calibration term introduced in the training loss on newly added classes, thereby ensuring balanced and robust classification performance for both new and old categories.
\item We conduct extensive experiments on image classification benchmarks, demonstrating that our method consistently reduces negative flips while maintaining or improving overall accuracy on the CIFAR100 and ImageNet1K datasets.
\end{itemize}

\section{Related Works}\label{sec:related}

\paragraph{Negative Flips Reduction.}

Negative flips, formalized by \cite{yan2021positive} expanding upon earlier notions of predictive churn \cite{milani2016launch, toneva2018empirical}, are related to the broader areas of continual learning \cite{chen2018lifelong, kirkpatrick2017overcoming}, incremental learning \cite{li2016learning, prabhu2020greedy}, and sequential learning \cite{DBLP:journals/corr/GoodfellowMDCB13, mccloskey1989catastrophic}. While these areas focus on reducing forgetting and maintaining average performance across tasks \cite{delange2021continual, kirkpatrick2017overcoming, lopez2017gradient, zenke2017continual}, negative flips instead highlight instance-level prediction instability.
To address this issue, most existing methods aim to align predictions across versions via ensemble techniques \cite{zhao2024elodi, xie2021regression} or knowledge distillation \cite{yan2021positive, parchami2024good, echterhoff2024muscle}. 
Distillation methods \cite{yan2021positive,parchami2024good,echterhoff2024muscle,trauble2021backward,jiang2021churn} typically focus on output alignment but do not explicitly preserve decision margins, which can increase negative flips \cite{wu2022generalized}. 
For instance, Positive-congruent training (PCT) \cite{yan2021positive} proposes Focal Distillation that selectively distills only correct predictions from the old model, reducing the occurrence of negative flips.
Ensemble methods reduce stochastic variations in training and improve class separation margins by averaging the outputs of multiple models \cite{zhao2024elodi, xie2021regression, bahri2021locally, zhang2021hot, cai2022measuring, li2023lightweight}, but incur a high cost at inference. To reduce deployment costs and mitigate negative flips, ELODI \cite{zhao2024elodi} distills the ensemble knowledge to a single model.
However, existing approaches either incur significant computational overhead or do not explicitly address the preservation of the decision margin when additional classes are introduced. For this reason, we propose a novel approach that overcomes both of these limitations.

\paragraph{Knowledge Distillation.}
Knowledge distillation \cite{hinton2015distilling,beyer2022knowledge} was originally developed to transfer knowledge from a larger teacher model to a smaller student. Since then, it has evolved to include advanced frameworks such as self-distillation \cite{zhang2019your,anil2018large, Caron_2021_ICCV}, teacher-ensemble strategies \cite{chebotar2016distilling,you2017learning,malinin2019ensemble,asif2020ensemble,stanton2021does}, continual learning \cite{li2016learning, asadi2023prototype, mistretta2024improving}, and compatibility-aware approaches \cite{shen2020towards, zhou2023bt, biondi2023cores, biondi2022cl2r, ricci2024backward, biondi2024stationary, ricci2025orthogonality}. Multi-source and multi-level distillation, leveraging several teachers or latent representations, have also been explored for greater robustness with applications in various contexts \cite{amirkhani2021robust, yuan2021reinforced, jiang2024mtkd, liu2020adaptive}.
While knowledge distillation has been widely adopted to reduce negative flips \cite{yan2021positive, zhao2024elodi}, existing approaches often rely on complex ensembles or single-base model distillation, which can be inefficient or limited, respectively.
We propose a double-source focal distillation framework combined with decision margin preservation, which matches ensemble performance using only two models and also overcomes the limitations of single-model distillation by better preserving decision boundaries and reducing negative flips.

\paragraph{Margin Preservation.}
The concept of margins is foundational in a wide range of machine learning algorithms \cite{bartlett1996valid, bartlett1998boosting, weinberger2009distance}. Both empirical and theoretical work show that margins are closely connected to neural network generalization \cite{bartlett2017spectrally, elsayed2018large, jiangpredicting, neyshaburrole}, and play a key role in tasks such as classification \cite{pleiss2020identifying}, face recognition \cite{deng2019arcface, DBLP:conf/cvpr/LiuWYLRS17, DBLP:conf/cvpr/WangWZJGZL018}, long-tail distribution learning \cite{ren2020balanced, cao2019learning, menonlong}, and continual learning \cite{bic, kirkpatrick2017overcoming}. Recent work demonstrates that the final stage of classifier training can be seen as a margin optimization problem, directly linking the number of classes to the geometry of the learned representation \cite{pmlr-v235-jiang24i}.
However, despite the recognized importance of margins for generalization and robustness, decision margin preservation has not been systematically studied in the context of negative flip reduction. Our work is the first to explicitly address this gap.

\section{Problem Formulation}\label{sec:neg_flip}

Given an image \( \mathbf{x} \in \mathcal{X} \) and its ground-truth label \( y \in \mathcal{Y} \), let \( q_{\phi} \) denote the predicted probability distribution over classes given an input for a model parameterized by \( \phi \). The final predicted class for an input \( \mathbf{x}_i \) is given by:
$$
\hat{y}_i = \operatorname*{arg\,max}_{y \in \mathcal{Y}} q_{\phi}(y \mid \mathbf{x}_i).
$$
Consider a base model \( \phi_{\textnormal{old}} \) trained on an initial set of classes \( \mathcal{Y}^{\text{old}} = \{1, 2, \dots, C^{\text{old}}\} \) and an improved version \( \phi_{\text{new}} \), which is trained on the expanded class set \( \mathcal{Y}^{\text{all}} = \mathcal{Y}^{\text{old}} \cup \mathcal{Y}^{\text{new}} \), where \( \mathcal{Y}^{\text{new}} = \{C^{\text{old}}+1, \dots, C^{\text{new}}\} \) represents the set of new classes. 
By analyzing the predictions of \( \phi_{\textnormal{old}} \) and \( \phi_{\text{new}} \) on the original class set \( \mathcal{Y}^{\text{old}} \), we can categorize different prediction outcomes based on \( \hat{y}^{\textnormal{old}} \) and \( \hat{y}^{\textnormal{new}} \).
Predictions are considered consistent when both models either correctly classify a sample (\( \hat{y}^{\textnormal{old}}_i = \hat{y}^{\textnormal{new}}_i = y_i \)) or misclassify it (\( \hat{y}^{\textnormal{old}}_i \neq y_i \land \hat{y}^{\textnormal{new}}_i \neq y_i \)). However, in some cases, the predictions differ. A positive flip occurs when a sample previously misclassified by \( \phi_{\textnormal{old}} \) is correctly classified by \( \phi_{\textnormal{new}} \).  
These flips are considered beneficial in model updates and should therefore be encouraged.
Conversely, a negative flip occurs when a sample that was correctly classified by \( \phi_{\textnormal{old}} \) is misclassified by \( \phi_{\textnormal{new}} \), making it a key factor in the phenomenon of performance regression. Figure~\ref{fig:prediction_flip} illustrates an example of both positive and negative flips.

\begin{table*}[t]
\centering
\setlength{\tabcolsep}{0.5em}
\resizebox{\linewidth}{!}{
\begin{tabular}{l
                cccc    %
                c       %
                cccc    %
                }
\toprule
\multirow{3}{*}{Method} 
    & \multicolumn{4}{c}{\textbf{CIFAR100}} && \multicolumn{4}{c}{\textbf{ImageNet1k}} \\
    \cmidrule{2-5} \cmidrule{7-10}
    & ER\lowbetter (\%) on $\mathcal{Y}^{\text{old}}$ 
    & ER\lowbetter (\%) on $\mathcal{Y}^{\text{all}}$ 
    & NFR\lowbetter (\%) 
    & Rel-NFR\lowbetter (\%) 
    && ER\lowbetter (\%) on $\mathcal{Y}^{\text{old}}$ 
    & ER\lowbetter (\%) on $\mathcal{Y}^{\text{all}}$ 
    & NFR\lowbetter (\%) 
    & Rel-NFR\lowbetter (\%) \\
\cmidrule{1-5} \cmidrule{7-10}
Old model 
    & 33.27 & - & - & - 
    && 25.55 & - & - & - \\
\cmidrule{1-5} \cmidrule{7-10}
No treatment
    & 40.40	&39.12 &14.04	&52.39
    && 28.40 & 32.88 & 9.15 & 43.30 \\
BCT~\cite{shen2020towards}
    & 41.28	&39.58	&14.88	&54.34
    && 27.90 & 33.17 & 8.99 & 43.28 \\
PCT-Naive~\cite{yan2021positive}
    & 	40.54	&40.32	&14.04	&52.20
    && 26.22 &33.17 &7.24	&37.06	\\
PCT-KL~\cite{yan2021positive}
    & 39.44	&38.79	&11.94	&45.63
    && 	27.73	&\underline{32.69} &8.09	&39.19\\
PCT-LM~\cite{yan2021positive}
    & 41.34	&41.83	&12.12	&44.19
    && 27.98	&34.70 &7.55	&36.24	 \\
ELODI~\cite{zhao2024elodi}
    & 37.98	&\underline{38.45}	&\underline{10.56}	&41.91
    && 27.28	&\textbf{32.13} &7.61	&37.46	 \\
ELODI$_{TopK}$~\cite{zhao2024elodi}
    & 39.06	&\textbf{38.32}	&12.46	&48.09
    && 28.51	&32.97 & 9.07 & 42.74 \\
\cmidrule{1-5} \cmidrule{7-10}
MPT-KL (Ours)
    & \textbf{34.32}	&38.55	&\textbf{9.26}	&\underline{40.67}
    && \textbf{24.68}	&32.76 &\underline{6.09}	&\underline{33.15}	 \\
MPT-LM (Ours)
    &	\underline{35.28}	&38.61	&\textbf{9.26}	&\textbf{39.57}
    && \underline{25.46}	&34.12 &\textbf{6.02}	&\textbf{31.76}	 \\
\bottomrule
\end{tabular}
}
\caption{Comparison of different methods on CIFAR100 and ImageNet1k with a ResNet-18 architecture. We report the error rate on old classes (ER on $\mathcal{Y}^{\text{old}}$), the overall error rate (ER on $\mathcal{Y}^{\text{all}}$), the Negative Flip Rate (NFR), and the Relative Negative Flip Rate (Rel-NFR).}
\label{tab:results_cifar100_embstandard}
\end{table*}

\section{Margin Preserving Training (MPT)}\label{sec:method}

\begin{figure}[t]
    \centering
    \includegraphics[width=0.9\linewidth]{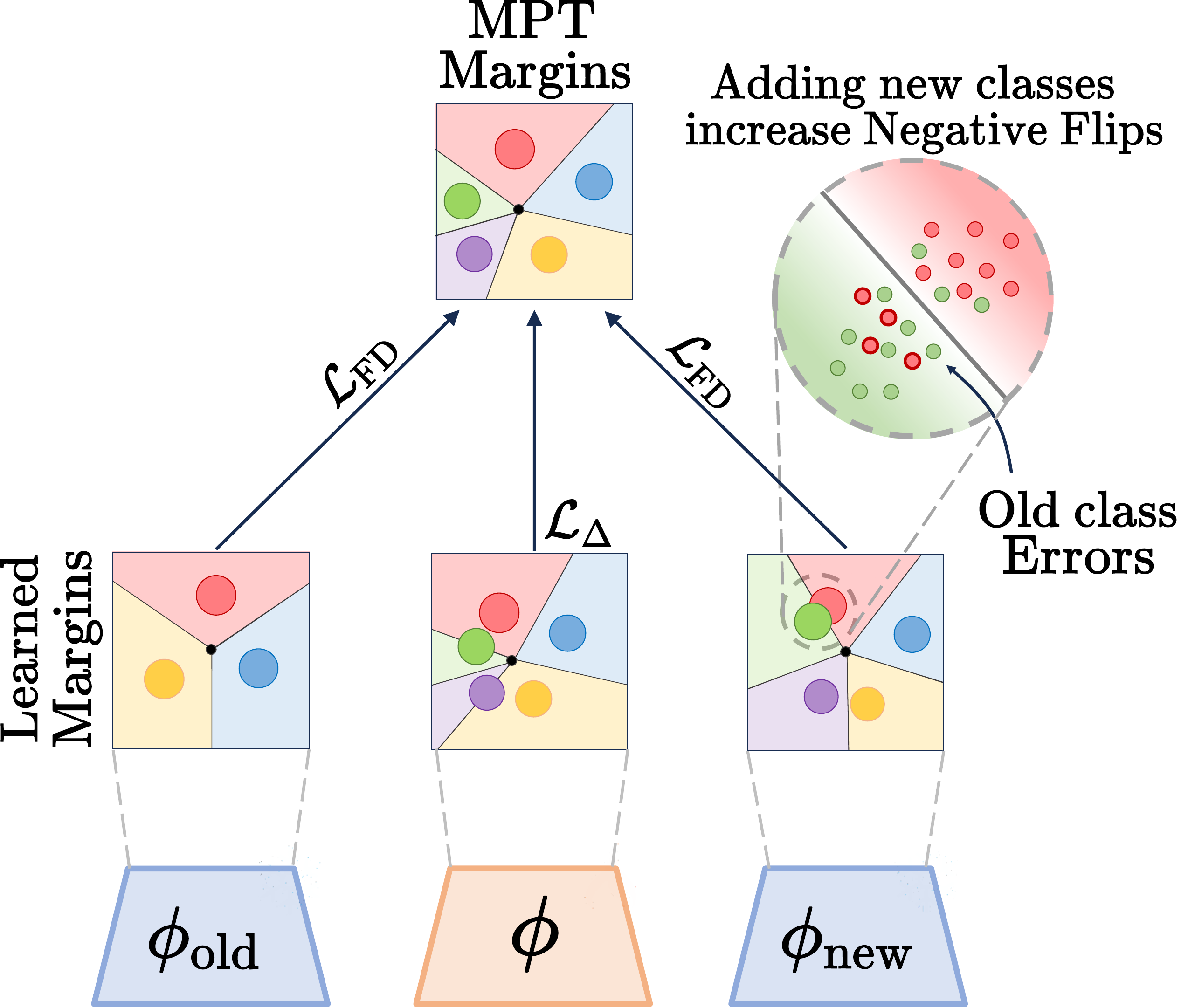}
    \caption{
    Schematic overview of the proposed Margin Preserving Training (MPT) approach. A model $\phi$ is trained with the margin-calibrated loss $\mathcal{L}_\Delta$ while receiving double-source focal distillation from two reference models, $\phi_\text{old}$ and $\phi_\text{new}$ (both optimized with cross-entropy). This combination mitigates margin underestimation for new classes, reduces negative flips between model updates, and preserves decision margins for old classes. The gray region marks areas near decision boundaries where negative flips typically occur. Margins are illustrated in a 2D embedding space for clarity, although MPT operates on logit-space margins over all classes.
    }
    \label{fig:method}
\end{figure}

In this section, we introduce our Margin Preserving Training method, which aims to reduce negative flips between successive model updates.
Our approach is characterized by (1) the incorporation of a bias term in the softmax cross-entropy loss (Equation~\ref{eqn:logit-adjusted-loss}) to preserve decision margins, and (2) the introduction of a double-source focal distillation term (Equation~\ref{eq:pcdouble}) to promote balanced learning between old and new classes.
Figure~\ref{fig:method} provides a schematic overview of our approach, illustrating the impact of each component on the learned class margins.

\subsection{Margin-calibrated Softmax Cross-entropy Loss} \label{sec:shift_logit}

When a model is updated by increasing the number of training classes, the margin between adjacent classes becomes narrower as they arrange themselves to maximize the minimum one-vs-rest margin \cite{pmlr-v235-jiang24i}. This contraction may lead to negative flips, particularly among instances of previously learned classes, as they become increasingly closer to each other. Furthermore, expanding the training set with new classes can introduce interfering features, thereby leading to negative transfer \cite{zhang2022survey}. This phenomenon arises because the model tends to learn more generic rather than class-specific features \cite{li2023identification}, which increases the probability of negative flips.
To address this, we propose a margin-calibrated softmax cross-entropy loss, designed to preserve the margins of previously learned classes. The core idea is to apply a positive logit bias to the newly introduced categories during training, tilting the decision boundary to favor of the old ones and thus maintaining their margin close to those learned by the base model, mitigating negative flips.

Given a logit vector $\mathbf{z}$ produced by a model $\phi$ for an input sample $\mathbf{x}$, where $\mathbf{z}_i$ corresponds to class $i$ (with $C$ the total number of classes), and $y$ is the ground-truth class label, we introduce a class-dependent logit bias $\Delta_i$.
The logit bias $\Delta_i$ is defined as a positive bias $k > 0$ applied exclusively to the newly introduced classes $\mathcal{Y}^{\text{new}}$:

\begin{equation}\label{eq:shift}
\Delta_i =
\begin{cases}
0, & \text{if } i \in \mathcal{Y}^{\text{old}},\\
k, & \text{if } i \in \mathcal{Y}^{\text{new}}.
\end{cases}
\end{equation}

The resulting margin-calibrated softmax cross-entropy loss is defined as:

\begin{equation}\label{eqn:logit-adjusted-loss}
\mathcal{L}_{\Delta} = -\log \frac{e^{ \mathbf{z}_y + \Delta_y}}{\sum_{i=1}^C e^{ \mathbf{z}_i + \Delta_i}}.
\end{equation}

Introducing a positive bias into the logit functions serves as an explicit prior, increasing the predicted confidence for targeted classes during training. This approach reduces the model's dependency on highly discriminative features for those classes, as the introduced bias inherently supports correct classification. Conversely, classes without the positive logit bias must rely exclusively on feature-based discrimination, prompting the model to develop larger decision margins for these unbiased categories. At inference time, with the bias removed, the model's confidence in previously biased classes diminishes, revealing their comparatively weaker internal representations. Consequently, the absence of this prior forces the model to maintain wider decision margins for unbiased categories, as these posed greater challenges during training without the benefit of a positive bias \cite{ren2020balanced}.
When the value of \(k=0 \), then \( \Delta_i = 0 \) for all classes, and Equation~\ref{eqn:logit-adjusted-loss} reduces to the standard cross-entropy loss, resulting in a balanced learning of margins across all categories.

Margin-calibrated losses have previously been formalized and employed to address class imbalance learning, where the most frequent classes receive larger margins due to disparities in class frequencies compared to rare ones~\cite{ren2020balanced, menonlong, cao2019learning}. Motivated by this, we reinterpret the class-dependent logit bias as a tool for more general margin regulation. In our approach, $\Delta_i$ is not determined by class frequencies, as is common in class imbalance methods, but is instead governed by a tunable hyperparameter $k$ that directly controls the trade-off between preserving old class margins and learning new classes.
Increasing the value of $k$ progressively biases the training in favor of previously learned classes, thereby allowing fine-grained control over the retention of old class margins.
While this strategy effectively reduces negative flips, it may also bias the model against new classes, leading to potential underfitting. To address this issue, we propose a double-source focal distillation mechanism in the following section.

\subsection{Double-source Focal Distillation Training}\label{sec:double_distil2}
When trained with the loss in Equation~\ref{eqn:logit-adjusted-loss} with a high value of $k$, the model \( \phi \) tends to preserve the decision margins of old classes but underestimates those of new classes due to the induced positive bias, resulting in significantly smaller decision margins for the latter and a higher error rate. In contrast, training a new independent model from scratch with standard cross-entropy (i.e., without the logit shift) results in more balanced margins and accuracy between old and new classes, but at the cost of increased negative flips.

\begin{table*}[t]
\centering
\setlength{\tabcolsep}{0.5em}
\resizebox{\linewidth}{!}{
\begin{tabular}{l
                cccc    %
                c       %
                cccc    %
                }
\toprule
\multirow{3}{*}{Method} 
    & \multicolumn{4}{c}{\textbf{CIFAR100}} && \multicolumn{4}{c}{\textbf{ImageNet1k}} \\
    \cmidrule{2-5} \cmidrule{7-10}
    & ER\lowbetter (\%) on $\mathcal{Y}^{\text{old}}$ 
    & ER\lowbetter (\%) on $\mathcal{Y}^{\text{all}}$ 
    & NFR\lowbetter (\%) 
    & Rel-NFR\lowbetter (\%) 
    && ER\lowbetter (\%) on $\mathcal{Y}^{\text{old}}$ 
    & ER\lowbetter (\%) on $\mathcal{Y}^{\text{all}}$ 
    & NFR\lowbetter (\%) 
    & Rel-NFR\lowbetter (\%) \\
\cmidrule{1-5} \cmidrule{7-10}
Old model 
    & 33.27 & - & - & - 
    && 25.55 & - & - & - \\
\cmidrule{1-5} \cmidrule{7-10}
No treatment
    & 37.26	& 36.33	& 12.20	& 49.36
    && 22.32	&26.38	&6.14	&36.92 \\
BCT~\cite{shen2020towards}
    & 39.08	& 37.34	& 13.58	& 52.38
    && 21.32	&25.34	&5.71	&35.98 \\
PCT-Naive~\cite{yan2021positive}
    & 37.38	& 37.96	& 12.12	& 48.88
    && 21.06	&26.25	&5.30	&33.78	\\
PCT-KL~\cite{yan2021positive}
    & 36.54	& 36.98	& 10.22	& 42.16
    && 		21.78	&26.03	&5.21	&32.14\\
PCT-LM~\cite{yan2021positive}
    & 39.20	& 41.02	& 10.50	& 40.38
    && 	24.16	&29.38	&5.15	&28.62	 \\
ELODI~\cite{zhao2024elodi}
    & 35.24	& 36.62	& \underline{9.10}	& \underline{39.45}
    && 20.42	&\textbf{24.36}	&4.27	&28.07	 \\
ELODI$_{TopK}$~\cite{zhao2024elodi}
    & 35.12	& \textbf{35.15}	& 10.14	& 43.52
    && 22.08	&26.16	&5.83	&35.46 \\
\cmidrule{1-5} \cmidrule{7-10}
MPT-KL (Ours)
    & \textbf{33.02}& \underline{36.07}	& \textbf{8.62}	&\textbf{39.35}
    && \textbf{19.23}	&\underline{24.95}	&\underline{3.79}	&\underline{26.48} \\
MPT-LM (Ours)
    & \underline{35.10}	& 37.28	& 9.42	& 40.46
    && \underline{19.84}	&25.50	&\textbf{3.49}	&\textbf{23.64} \\
\bottomrule
\end{tabular}
}
\caption{Comparison of different methods on CIFAR100 and ImageNet1k when also the architecture of the new model is updated from a ResNet-18 to a ResNet-50. We report the error rate on old classes (ER on $\mathcal{Y}^{\text{old}}$), the overall error rate (ER on $\mathcal{Y}^{\text{all}}$), the Negative Flip Rate (NFR), and the Relative Negative Flip Rate (Rel-NFR).}
\label{tab:results_cifar100_embstandard_resnet50}
\end{table*}

To address this trade-off, we propose a double-source focal distillation strategy (see Figure~\ref{fig:method}). This approach transfers positive knowledge from both the old model $\phi_{\textnormal{old}}$ and a new reference model $\phi_{\textnormal{new}}$, trained on all categories with standard cross-entropy loss, which provides reliable guidance also for the newly added classes. By combining both sources, the final model maintains robustness on old classes while improving classification on new ones, even when subjected to an induced positive bias.
The proposed final training objective is:
\begin{equation}\label{eq:pcdouble}
    \min_{\phi} \mathcal{L}_{\Delta} + \lambda \mathcal{L}_{\textnormal{FD}}(\phi, \phi_{\textnormal{old}}) + \lambda \mathcal{L}_{\textnormal{FD}}(\phi, \phi_{\textnormal{new}})
\end{equation}
where $\mathcal{L}_{\Delta}$ is the logit-adjusted softmax cross-entropy loss from Equation~\ref{eqn:logit-adjusted-loss}, and $\mathcal{L}_{\textnormal{FD}}$ a focal distillation loss.
We adopt the Focal Loss distillation terms as proposed by~\citet{yan2021positive} to better emphasize instances correctly classified by the reference model and mitigate the risk of suppressing positive flips. The formal definition of $\mathcal{L}_{\textnormal{FD}}$ is given by:
\begin{equation}\label{eq:distll}
    \mathcal{L}_{\textnormal{FD}}= \sum_{i = 1}^{N} [\alpha + \beta \cdot \mathbf{1}(\hat{y}^*_i = y_i)] \cdot \rm d(\phi, \phi^*),
\end{equation}
where $\alpha$ is a base weight for all samples, $\beta$ is an additional weight for samples correctly predicted by a reference model $\phi^*$, and $\rm d$ is the distance between model outputs.
In practice, the distance $\rm d$ can be defined as a temperature-scaled Kullback-Leibler (KL) divergence~\cite{hinton2015distilling}:
\begin{equation}\label{eq:KD}
\rm d_{\text{KL}}(\phi_1, \phi_2) = \text{KL} \left( \sigma \left(\frac{\mathbf{z}_1}{\tau} \right), \sigma \left(\frac{ \mathbf{z}_2}{\tau} \right) \right)
\end{equation}
where $\sigma$ is the softmax, $\tau$ is the temperature, and $\mathbf{z} = \phi(\mathbf{x})$ the logit vector for input $\textbf{x}$. Alternatively, the Euclidean distance applied directly to the logits may be used~\cite{hinton2015distilling, bucilua2006model}:
\begin{equation}
\rm d_{\text{LM}}(\phi_1, \phi_2) = \frac{1}{2} ||\mathbf{z}_1 - \mathbf{z}_2||_2^2.
\label{eq:logit_matching}
\end{equation}

In the following, we refer to this approach as Margin Preserving Training (MPT).

\section{Experimental Results}\label{sec:experiments}

\subsection{Experimental Setup}\label{sec:ecperimental_setup}

\paragraph{Datasets.}
We evaluate the proposed approach using two standard image classification datasets: CIFAR100 \cite{Krizhevsky2009LearningML} and ImageNet1K \cite{russakovsky2015imagenet}.
ImageNet1K contains 1,000 categories with 1,281,167 training images and 50,000 test images. CIFAR100 comprises 100 categories, including 50,000 training and 10,000 test images. 
Additionally, we employ CIFAR10 \cite{Krizhevsky2009LearningML} to present qualitative results.
For data splits, we follow the methodology from \cite{yan2021positive}, using 50\% of classes ($\mathcal{Y}^\textnormal{old}$) for training the old model and all classes ($\mathcal{Y}^\textnormal{all}$) for training the updated model.

\paragraph{Implementation Details.}
For CIFAR100 and CIFAR10, we train a ResNet-18 and a ResNet-50~\cite{he2016resnet} using SGD (momentum = 0.9) with a base learning rate of 0.1, applying cosine annealing \cite{loshchilov2016sgdr} over 100 epochs with a batch size of 128.
For ImageNet1k, we adopt the standard ImageNet1k training recipe provided directly by PyTorch repository. Specifically, we train a ResNet-18 and a ResNet-50~\cite{he2016resnet} with embedding dimension 128 using SGD (momentum = 0.9) with an initial learning rate of 0.1, decayed by a factor of ten every 30 epochs for a total of 90 epochs. The batch size is set to 128. For our method, we set $\lambda=0.4$ and $\lambda=1.0$ when LM of Equation \ref{eq:logit_matching} and KL of Equation \ref{eq:KD} are used as distillation functions, respectively. 
Following prior work \cite{yan2021positive, zhao2024elodi}, we train every model from scratch, as this represents the most challenging setting, resulting in the highest NFR \cite{yan2021positive}.

\begin{figure*}
    \centering
    \includegraphics[width=\linewidth]{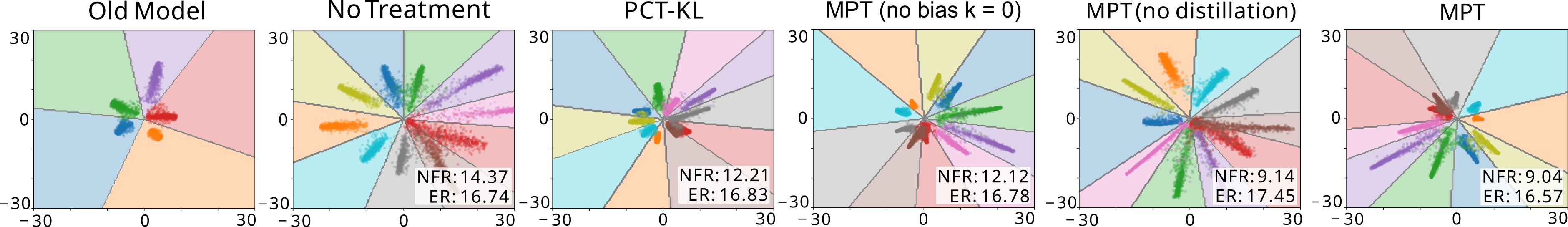}
    \caption{Comparison of embedding spaces for ResNet-18 models (embedding size = 2) on the CIFAR10 test set under various update strategies. Notably, when the model is trained directly on multiple classes (``No Treatment''), the inter-class margins are narrower compared to the original model. Our proposed Margin Preservation Training (MPT) maintains the margins of the original model, thereby significantly reducing the rate of negative flips without compromising the error rate across all classes.}
    \label{fig:cifar10_2D_qualitative}
\end{figure*}

\paragraph{Evaluation Metrics.}
To measure model accuracy, we report the error rate (ER) computed on the full test set ($\mathcal{Y}^\textnormal{all}$). 
Additionally, we report ER specifically on test samples from the old classes ($\mathcal{Y}^\textnormal{old}$), enabling comparison of the accuracy between base and updated models. As minimizing negative prediction flips (Figure \ref{fig:prediction_flip}) is the main goal of our approach, we adopt the Negative Flip Rate (NFR) \cite{yan2021positive}, defined as:
\begin{equation}
\textnormal{NFR} = \dfrac{1}{N} \sum_{i = 1}^{N} \mathbf{1}(\hat{y}_{i}^{\text{new}} \neq y_i \land \hat{y}_{i}^{\text{old}} = y_i),
\end{equation}
where $N$ is the number of test samples, $\mathbf{1}(\cdot)$ is the indicator function, $\hat{y}_{i}^{\text{new}}$ and $\hat{y}_{i}^{\text{old}}$ denote predictions of the new and old models respectively, and $y_i$ is the ground truth label. Since NFR alone does not account for differing model accuracies, we also report the Relative Negative Flip Rate (Rel-NFR) \cite{yan2021positive}, computed as:
\begin{equation}
\textnormal{Rel-NFR} = \frac{\textnormal{NFR}}{(1 - \textnormal{ER}_{\textnormal{old}}) \cdot \textnormal{ER}_{\textnormal{new}}},
\label{eq:rel_nfr}\end{equation}
which normalizes negative flips relative to model accuracy differences.
The NFR metric measures only cases where previously correct predictions become incorrect. As such, it can decrease even if the overall error on old classes increases slightly. For instance, some examples that were
previously misclassified may be correctly classified by the updated model.

\paragraph{Compared Methods.}
The “Old model” refers to a baseline trained solely on the original dataset (i.e., 50\% of the classes). “No Treatment” denotes a setting in which the model is updated using the standard cross-entropy loss, without explicit mechanisms to mitigate negative flips. For comparison, we also include the PCT-Naive approach from \cite{yan2021positive}, as well as variants of PCT that employ alternative focal distillation losses: KL divergence (PCT-KL, Equation~\ref{eq:KD}) and Logit Matching (PCT-LM, Equation~\ref{eq:logit_matching}).
ELODI~\cite{zhao2024elodi} is implemented as logit distillation from an ensemble of eight independently trained new models and a single old model. Its TopK variant, which distills only from the Top-K components of the logits, is also included as a baseline (we set K=10 as in \citet{zhao2024elodi}).

\subsection{Quantitative Results}\label{sec:main_results} 

We present quantitative comparisons of our proposed MPT methods (MPT-KL and MPT-LM) against existing baselines in Tables~\ref{tab:results_cifar100_embstandard}, where the same architecture is used for both the old and new models, and~\ref{tab:results_cifar100_embstandard_resnet50}, where the new model adopts a more powerful architecture (ResNet-50) than the original (ResNet-18). The results show that both MPT variants achieve competitive error rates (ER) on old and new classes, indicating effective retention of prior knowledge. Notably, both methods significantly reduce the NFR, highlighting their ability to preserve correct predictions made by the original model.
In both experimental settings, MPT-KL and MPT-LM consistently achieve lower NFR values than all baseline methods, notably outperforming ELODI, which employs an ensemble of multiple new models. These findings suggest that preserving the decision margins of the original model plays a key role in mitigating negative flips.
The Rel-NFR metric further corroborates this observation: MPT-KL and MPT-LM consistently outperform all baselines, even when controlling for overall accuracy. This indicates that MPT maintains prior knowledge without simply sacrificing accuracy on new classes. Overall, these results clearly demonstrate the effectiveness of the MPT approach in reducing negative flips while preserving performance on previously learned classes.

To further validate our approach on modern architectures, we conduct experiments by fine-tuning only the classifier layer of a ViT-B/32 model pretrained on ImageNet-1k, using the CIFAR100 dataset. As shown in Table~\ref{tab:vit}, MPT achieves the best results in terms of error rate for both old and new classes, as well as the lowest value of NFR, thereby validating the effectiveness of our approach.

\begin{table}[t]
\centering
\setlength{\tabcolsep}{0.6em}
\resizebox{0.85\linewidth}{!}{
\begin{tabular}{cccccc    %
                }
\toprule
Bias & $\rm d$
    & ER$(\mathcal{Y}^\text{old})$ & ER$(\mathcal{Y}^\text{all})$ & NFR & Rel-NFR \\
\midrule
$\times$ & $\times$  
    & 28.40	&32.88  &9.16	&43.31	 \\
$\times$ & LM 
    & 27.98	&34.70 &7.55	&36.24	 \\
$\times$ & KL  
    & 27.73	&\textbf{32.69} &8.09	&39.19	 \\
\checkmark & $\times$  
    & 	26.23	&34.13 &7.71	&39.49\\
\checkmark & KL 
    & 	\textbf{24.68}	&\underline{32.76} &\underline{6.09}	&\underline{33.15}	 \\
\checkmark & LM 
    & \underline{25.46}	&34.12 &\textbf{6.02}	&\textbf{31.76}	 \\
\bottomrule
\end{tabular}
}
\caption{
Ablation study of design choices of MPT on ImageNet1k. We report the error rate on old classes 
(ER$(\mathcal{Y}^\text{old})$), the overall error rate (ER$(\mathcal{Y}^\text{all})$), the Negative Flip Rate (NFR), and the Relative Negative Flip Rate (Rel-NFR). Variants remove margin-based regularization or distillation. The last two rows report performance of our method, i.e., MPT-KL and MPT-LM.
}
\label{tab:ablation_merged}
\end{table}

\subsection{Qualitative Results}
\label{sec:qualitative_results}
To complement the quantitative analysis presented in previus section, we provide qualitative results obtained from training a ResNet-18 model with a two-dimensional feature space on CIFAR10 using different methods.  
A two-dimensional feature space enables direct visualization of the learned representations without the need for dimensionality reduction techniques such as t-SNE~\cite{van2008visualizing} or UMAP~\cite{mcinnes2018umap}. Figure~\ref{fig:cifar10_2D_qualitative} presents the resulting representations across six different settings: (1) ``Old Model'', (2) ``No Treatment'', (3) PCT-KL, (4) MPT (no bias, $k=0$), (5) MPT (no distillation, $\lambda = 0$ in Equation \ref{eq:pcdouble}), and (6) MPT. In each subplot, data points are color-coded according to their true class labels.  

In the "Old Model" each class forms a compact cluster centered in its angular sector and is well-separated from adjacent decision boundaries. In "No Treatment" these clusters drift outward and increasingly intersect neighboring sectors: the green and blue points cross their boundaries, as do the red and newly introduced brown classes. This indicates substantial margin shrinkage and a tighter geometric arrangement relative to the old model, consistent with the findings of \cite{wu2022generalized}.
The PCT-KL method attempts to replicate the feature distribution of the Old Model, resulting in clusters that are more concentrated near the origin compared to the "No Treatment" scenario. However, not all old class margins are well preserved: some red class points are within the brown class region, indicating a uniform reduction of margins for that class.
MPT (no bias, $k=0$) produces effects similar to those of PCT-KL. By leveraging knowledge from the new model, it prevents new class clusters from collapsing toward the center or into old class regions, resulting in a slightly lower error rate.
MPT (no distillation) clearly preserves the margins of the old classes, as all old class points remain within their respective regions. However, this approach results in a high error rate for the new classes, as they are mapped into regions corresponding to the old classes, thereby yielding large decision margins for the new classes.
Our proposed approach, MPT, achieves the best results.
It provides superior margin preservation for old classes and a lower error rate for new classes, with each point correctly residing in its respective class region. 
Figure~\ref{fig:cifar10_2D_qualitative} shows the emergence of interfering features with the introduction of new categories, as discussed by \citet{tirer2023perturbation}. For instance, the new brown class (Dog) leads to interference in the learning of the old red class (Cat), as they are semantically similar.

\subsection{Ablation Studies}\label{sec:ablation}

\begin{figure}
    \centering
    \includegraphics[width=0.9\linewidth]{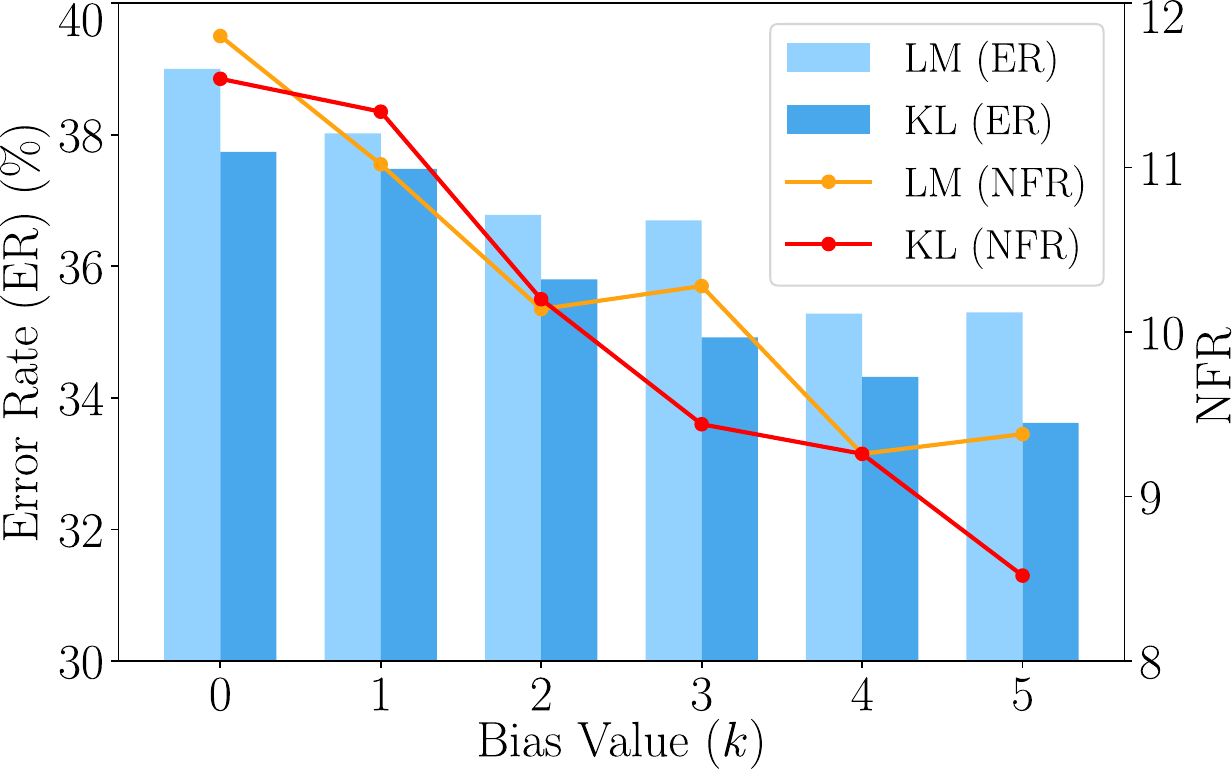}
    \caption{Effect of different margin value $k$ of Equation \ref{eq:shift} on CIFAR100 for both LM and KL distances in the focal distillation objectives. Bars show the error rate (ER), and lines represent the negative flip rate (NFR).}
    \label{fig:ablation_bias_values}
\end{figure}

\paragraph{Design Choices of MPT.}
To better understand the contributions of each component in our method, we analize the effects of margin-based regularization and double-source focal distillation. Table~\ref{tab:ablation_merged} shows the impact of removing margin-based regularization ($k=0$ in Equation~\ref{eq:shift}) or double-source focal distillation by setting $\lambda = 0$ in Equation~\ref{eq:pcdouble}.
The results show that both margin-based regularization and focal distillation are essential for minimizing negative flips. Removing either component increases the Negative Flip Rate (NFR) and Relative Negative Flip Rate (Rel-NFR), highlighting their complementary effects. Margin-based regularization alone improves performance on old classes but reduces accuracy on new classes, while distillation alone increases negative flips.
Our proposed methods, MPT-KL and MPT-LM, which combine both components, consistently achieve superior performance compared to variants using only one. This confirms that integrating margin preservation and focal distillation is crucial for balanced performance, minimizing negative flips, and maintaining high accuracy across both old and new classes.

\paragraph{Exploratory Studies of Parameters in MPT.}
We evaluate the effect of varying the margin parameter $k$ (Equation~\ref{eq:shift}) on the error rate (ER) and Negative Flip Rate (NFR) in CIFAR100, as shown in Figure~\ref{fig:ablation_bias_values}.
Owing to computational constraints, these experiments were not repeated on ImageNet-1k. 
Figure~\ref{fig:ablation_bias_values} shows that increasing $k$ generally decreases negative flips while maintaining a competitive error rate, with $k=4$ providing the optimal trade-off for CIFAR100.
This trend holds for both KL-based and LM-based distillation objectives.
Consequently, we set $k=4$ for all CIFAR100 experiments.
For ImageNet-1k, a smaller margin ($k=1.5$) is used to better suit the larger dataset.

\begin{table}[t]
	\centering
	\setlength{\tabcolsep}{0.5em}
	\resizebox{0.95\linewidth}{!}{
		\begin{tabular}{l c c c c}
			\toprule
			\multirow{2}{*}{Method} & \multicolumn{2}{c}{ER\lowbetter (\%) on} & NFR\lowbetter (\%) & Rel-NFR\lowbetter(\%) \\
            \cmidrule{2-3}
			& $\mathcal{Y}^{\text{old}}$ & $\mathcal{Y}^{\text{all}}$ & &  \\
			\cmidrule{1-5}
			Old model & 13.28 & - & - & - \\
			\cmidrule{1-5}
			No treatment  & 19.44 & 19.60 & 7.08 & 41.86   \\
            BCT~\cite{shen2020towards}  & 19.78 & 19.89 & 6.64 & 38.70\\
			PCT-Naive~\cite{yan2021positive} & 17.40 & 19.73 & 5.26 & 34.85 \\
			PCT-KL~\cite{yan2021positive} & 19.44 & \underline{19.52} & 6.88 & 40.81 \\
			PCT-LM~\cite{yan2021positive} & 19.50 & 19.59 & 6.58 & 38.91 \\
            ELODI~\cite{zhao2024elodi} & 17.13 & 19.54 & 4.97 & 30.69 \\
            ELODI$_{TopK}$~\cite{zhao2024elodi} & 17.52 & 19.57 & 5.16 & 31.98 \\
			\cmidrule{1-5}
			MPT-KL (Ours) & \textbf{15.16} &  \textbf{19.51} & \textbf{3.26} & \textbf{24.09} \\
			MPT-LM (Ours) & \underline{16.22} & 19.58 &  \underline{3.58} &  \underline{25.45} \\
			\bottomrule
		\end{tabular}
	}
	\caption{Negative Flip Reduction comparison of a pretrained ViT-B/32  fine-tuned on CIFAR100.}
	\label{tab:vit}
\end{table}

\section{Conclusions}\label{sec:conclusions} 
In this paper, we investigated the problem of negative flips, where updated models misclassify samples previously classified correctly. 
We addressed this issue from a novel margin-based perspective by explicitly preserving decision margins learned by the old model during incremental updates. 
Specifically, we proposed Margin Preserving Training (MPT), which applies a logit bias to maintain decision margins for previously learned classes and integrates additional knowledge via a double-source focal distillation strategy. 
Extensive experiments on CIFAR100 and ImageNet-1k demonstrate that MPT significantly reduces negative flips while maintaining competitive accuracy across all classes. 

\paragraph{Limitations.} Our method requires training an additional reference model on all classes, modestly increasing computational demands. However, this overhead remains lower than typical ensemble-based methods. MPT's performance depends on tuning of the margin bias, emphasizing the need for empirical optimization across datasets and scenarios.

\section*{Acknowledgments}
This paper was partially funded by the project "Collaborative Explainable neuro-symbolic AI for Decision Support Assistant", CAI4DSA, CUP B13C23005640006.

\bibliography{aaai2026}

\end{document}